\documentclass{article}
\usepackage{graphicx} % Required for inserting images
\usepackage{hyperref}
\usepackage{enumitem}
\usepackage[textheight=8in]{geometry}
\usepackage{tcolorbox}
\usepackage{array}
\usepackage[sort]{cite}

\title{Artificial Intelligent Disobedience: \\ Rethinking the Agency of Our Artificial Teammates}
\author{Reuth Mirsky}
\date{}

\begin{document}

\maketitle

\begin{abstract}
Artificial intelligence has made remarkable strides in recent years, achieving superhuman performance across a wide range of tasks. Yet despite these advances, most cooperative AI systems remain rigidly obedient, designed to follow human instructions without question and conform to user expectations, even when doing so may be counterproductive or unsafe. This paper argues for expanding the agency of AI teammates to include \textit{intelligent disobedience}, empowering them to make meaningful and autonomous contributions within human-AI teams. It introduces a scale of AI agency levels and uses representative examples to highlight the importance and growing necessity of treating AI autonomy as an independent research focus in cooperative settings. The paper then explores how intelligent disobedience manifests across different autonomy levels and concludes by proposing initial boundaries and considerations for studying disobedience as a core capability of artificial agents.
\end{abstract}

\section{Introduction}
The field of artificial intelligence is currently abuzz with discussions surrounding ``agentic AI'' or ``AI agents.'' However, despite the widespread excitement, the term \textit{agent} itself often lacks a precise, universally agreed-upon definition within these conversations. Recently, significant focus has shifted towards agents built upon large language models (LLMs), leveraging some reasoning and language understanding capabilities to execute complex tasks, interact with external tools, and learn from feedback \cite{wei2022chain, yao2023react, schick2023toolformer, yao2023tree, shinn2023reflexion}. This move towards more autonomous, goal-directed LLM systems represents a promising yet challenging frontier in AI development. During this time, AI algorithms have also reached superhuman performance in numerous tasks such as game playing \cite{brown2019superhuman, silver2017mastering, vinyals2019grandmaster, wurman2022outracing} and text and image processing \cite{antoniou2017data, devlin2018bert, saharia2022photorealistic}. On the other hand, there are still significant obstacles that modern AI has yet to overcome. Grosz \cite{grosz2012question} proposed a revised Turing Test to create:

\begin{quote} "\textit{A computer team member that can behave, over the long term and in uncertain, dynamic environments, in such a way that people on the team will not notice that it is not human}." \end{quote}

This vision captures not just the importance of task performance, but also fluency, trust, and appropriate initiative in collaborative settings. This work argues that such agency must include the ability to make context-sensitive decisions that may sometimes involve \textit{intelligent disobedience}.

In most existing work on cooperative AI agents and robots, it is assumed that artificial agents should do their best to follow the instructions they are given \cite{christian2020alignment}. However, as AI agents achieve human-comparable and even superhuman performance in various domains, there is a growing need to reason more deeply about the levels of autonomy appropriate for such agents \cite{coman2018ai}. This paper introduces and analyzes a taxonomy of autonomy levels, ranging from zero autonomy (pure obedience) to full autonomy, where the agent may independently generate and revise goals. Each level corresponds to increasing decision-making power and situational flexibility.

\textbf{This paper calls on AI practitioners to extend the agency of AI teammates beyond strict obedience, allowing them to act with increased autonomy when necessary.} Consider a guide dog, trained to intelligently disobey: if it is given a clearly unsafe command from its handler, it is taught to disobey it \cite{mirsky2021seeing}. The paper will use this example and others to showcase how we might reconsider the agency of artificial teammates, so they can be better collaborators.

This paper further contributes to a formalization of intelligent disobedience as a critical capacity in autonomous agents. It identifies levels of disobedience corresponding to the agent’s autonomy level. These levels delineate when, how, and to what extent an agent should override instructions, ranging from reflexive overrides (e.g., avoiding immediate harm) to sophisticated mediations that involve alternative plan generation and moral reasoning.

This paper argues that investigating AI agency in cooperative settings as an independent capability should be part of the modern AI research agenda for several reasons. First, it challenges contemporary perceptions regarding the autonomy that should be given to AI agents and robots, especially in situations where such intelligence outperforms human judgment. Changing the autonomy of AI in such cases has the potential to add significant value to the team by making the AI a uniquely contributing member. Second, AI agency forces discussion on the boundaries and restrictions of autonomous AI, a necessary topic as AI capabilities continue to expand. As technology progresses and AI systems gain greater autonomy, regulations enabling AI to make decisions must be updated \cite{PauseAI2023, middleton2022trust}. As insulin pumps, autonomous vehicles, and surgical robots gain more autonomy to act without human intervention, it becomes crucial to ensure that these systems act intelligently and appropriately in their context.

Finally, this work explores the boundaries of disobedience: when disobedience is warranted, when it becomes a liability, and how to design agents that can transparently mediate between global safety principles and local human instructions. Using frameworks like the five-step model proposed by Mirsky and Stone \cite{mirsky2021seeing}, this work situates intelligent disobedience within broader efforts in AI alignment and safety. The goal of this work is to equip artificial agents not only with competence but with the moral and contextual sensitivity needed to be truly effective teammates.

\section{Defining AI Agency}

To start discussing AI agency, this paper follows the definition of AI by Nilsson \cite{nilsson2009quest}: 
\begin{quote}
``\textit{Artificial intelligence is that activity devoted to making machines intelligent, and intelligence is that quality that enables an entity to function appropriately and with foresight in its environment}''.
\end{quote}
What ``to function appropriately'' entails highly depends on context. 
Generally, it refers to systems exhibiting autonomous behavior, capable of (1) perceiving their environment, (2) making decisions, and (3) taking actions to achieve specific goals \cite{russell2016artificial,wooldridge2009introduction}. Moreover, creating truly effective AI agents necessitates equipping them with \textit{persistent memory} that carries context and state across separate interactions, preventing each call from being a disconnected event. This technical requirement echoes deep philosophical considerations about identity and memory that are beyond the scope of this work  \cite{locke1847essay, tulving1985memory}. Out of this discussion, we need to retain that an agent has a \textit{persistent mission}: a goal or a set of objectives it is inherently meant to achieve.

This persistent mission is a crucial capability of appropriately functioning collaborative AI systems: A guide robot must not let a visually impaired person cross a road when a car is approaching; A medical management software should alert a physician who prescribes a patient a seemingly wrong dosage of medicine; and for a service robot, it might be preferable to interrupt its owner and ask for clarifications instead of executing a wrong task. These examples show that AI agency is more than a set of restrictions but rather a \textbf{deliberation process} that can lead to different outcomes under different circumstances.
Taking the five levels of vehicle autonomy, this paper expands the scale system to artificial autonomy, particularly focusing on the ability to make decisions independently of a human teammate. At all autonomy levels above 0, the agent is assumed to be adhering to its persistent mission and to stay committed to cooperating with its teammates \cite{grosz1999evolution}:
\begin{tcolorbox}[title=Autonomy Levels of an Artificial Agent, colback=blue!5!white, colframe=blue!75!black, fonttitle=\bfseries]

\begin{enumerate}[label=\textbf{L\arabic*}, start=0]
    \item \textit{No autonomy} --- parallel to hard-coding a task. For example, a factory robot arm is programmed to follow a fixed sequence of motions to assemble parts, without sensing or decision-making capabilities. If the object is misaligned or missing, it continues regardless, potentially failing the task.
    
    \item \textit{Support} --- support actions that the human performs. For example, a grammar-checking tool that suggests corrections as a user types, but doesn’t change anything unless the user approves.
    
    \item \textit{Occasional Autonomy} --- acts autonomously on well-defined sub-tasks. For example, a robotic vacuum that maps a room and autonomously cleans it, but needs human input to avoid new obstacles or to resume after a pause.
    
    \item \textit{Limited autonomy} --- full autonomy in some conditions, but might ask the human to take over. Beyond the previous level of autonomy, agents at this level have the autonomy to decide when to transfer control back to a human. Such an agent can be a surgical robot that can autonomously suture a wound during a procedure, but calls back to a human surgeon to handle unexpected bleeding or any deviation from the standard procedure.
    
    \item \textit{Full autonomy under constraints} --- The agent has a continuous full autonomy, which the human can override. For example, a warehouse robot that autonomously navigates and manages inventory within a fenced facility, but allows supervisors to override or redirect it if needed.
    
    \item \textit{Full autonomy} --- The agent has full autonomy in all conditions. For example, an AI-powered digital agronomist that continuously collects and interprets data from satellites, IoT sensors, and weather forecasts to make decisions on planting schedules, irrigation, fertilization, pest management, and harvesting. It communicates directly with automated machinery and supply chains, adapts to new environmental conditions, and revises long-term strategies, without requiring human oversight.
\end{enumerate}

\end{tcolorbox}

 This is not the first work that aims to expand and detail the various levels of autonomy for agents beyond autonomous vehicles. Beer et al. \cite{beer2014toward} presented the 10 Levels of Robot Autonomy (LORA): manual control, action support, batch processing, shared control, decision support, blended decision making, rigid system, automated decision making, supervisory control, and full automation. While most of these levels can be clearly mapped into the scale presented here, they are meant mainly to capture various modes of collaboration and shared responsibility along a continuum, focusing more on the allocation of functions (Sense, Plan, Act) between the human and the robot. 
Lopes et al. \cite{lopes2024exploring} similarly emphasizes domain-specific applications (e.g., diagnosis support, monitoring, or decision-making) and relates autonomy to data adaptivity and workflow integration rather than a structured control hierarchy. Zhang et al. \cite {zhang2023follower} explore autonomy from the perspective of user roles in guidance tasks, especially for visually impaired users, blending social interaction with control-sharing, which blurs the boundaries between levels like L1 (support) and L2–L3 (partial autonomy) in the scale presented here. Olatunji et al. \cite{olatunji2021levels} suggest a model for integrating LORA and levels of transparency (LoTs) in assistive robots to match the preferences and expectations of older adults, suggesting a co-adaptive design approach rather than a strictly hierarchical one. Finally, Milliken and Milliken \cite{milliken2017modeling} propose a dynamic Partially Observable Markov Decision Model (POMDP) where the agent selects autonomy levels based on inferred user expertise, where the level of control given between the user and the robot is encapsulated in macro-action controllers. Their approach is orthogonal to the scale presented here and could be used to implement various autonomy levels.

\section{Agents that Intelligently Disobey an Instruction}
Once full autonomy is achieved, the need for a human-in-the-loop diminishes. However, even before reaching this stage, there may be instances where the machine’s decision-making surpasses that of its human counterpart. This situation is common in many teams, and a good teammate often contributes unique capabilities to the team. A clear example of such behavior can be seen in modern grammar checkers. These tools often understand grammatical rules better than most users, and they now frequently autocorrect errors without requiring explicit permission from the user, even though their autonomy is limited to support (L1 autonomy). This phenomenon raises an important question: when should we allow autonomous agents not only to make decisions independently, but also to \textit{override} the instructions of human teammates?

Early work on \textit{command rejection} provides a starting point for answering this question. Briggs and Scheutz \cite{briggs2015sorry} presented a decision procedure for evaluating a human directive and demonstrated how a robot could explain its refusal. Their account already considered competing goals, social obligations, and the anticipated consequences of compliance. Briggs et al. \cite{briggs2022and} developed this account further, examining how a robot can recognize a norm violation and communicate its objection. As Arnold, Briggs, and Scheutz \cite{arnold2022only} argue, disobedience also presupposes that the agent understands the instruction it departs from. Command rejection concerns the immediate decision of whether to carry out a particular directive. Intelligent disobedience, as used here, extends that reasoning to the plan the directive serves and its implications for the agent’s persistent mission. A guide robot, for example, might safely take the next turn its handler requests yet recognize that the route will eventually lead to a crossing the handler cannot safely navigate. By proposing a different route now, the robot can preserve the handler’s goal of reaching the destination while fulfilling its standing commitment to safe mobility.

Following this thought process, Mirsky and Stone \cite{mirsky2021seeing} propose a structured framework for intelligent disobedience in autonomous agents, particularly within assistive contexts like guide robots. Existing autonomous systems enforce safety through hard constraints upon the system's abilities. Designing intelligent disobedient agents that ensure safety is a more intricate challenge. The main contribution of many modern systems is to provide a better understanding of the instructions given by a human \cite{hadfield2016cooperative}, within the limits of those safety restrictions \cite{kapoor2006constrained, rozo2013learning, agarwal2015impedance, lankenau1998safety}. Moreover, modern AI systems need to reason about cases in which the ``right'' thing to do is the opposite of the instruction given by the handler.
The \textit{intelligent disobedience model} outlines five sequential cognitive steps that an agent must undertake to safely and effectively override human commands when necessary:

\begin{enumerate}
    \item Global Objectives: The agent must internalize overarching, standing goals such as ``keep the handler safe.'' This step requires a comprehensive understanding of the environment and the agent's capabilities, ensuring adherence to fundamental safety constraints.
    \item Local Objectives: The agent needs to interpret the handler's immediate intentions or commands, which may change with time \cite{shamirodgr,elhadad2025,shamir2025}. 
    \item Plan Recognition: The agent must infer the handler's intended plan on \textit{how} to achieve the local objectives. This step necessitates a theory of mind, allowing the agent to anticipate the handler's actions based on environmental context and perceived goals.
    \item Consistency Check: Here, the agent evaluates whether executing the inferred plan aligns with the local and global objectives. If a conflict is detected, such as a command that could lead to potential harm, the agent recognizes the need for intervention.
    \item Mediation: Upon identifying a conflict, the agent deliberates on the appropriate course of action. This step could involve proposing alternative plans, seeking clarification, or, if necessary, overriding the command to prevent harm. The mediation process strikes a balance between respecting human autonomy and upholding safety and ethical standards.
\end{enumerate}

This framework highlights intelligent disobedience as a nuanced, context-sensitive process, enabling agents to function as collaborative partners rather than mere executors of commands. To preserve the agent’s commitment to its persistent mission and user expectations, disobedience should be appropriate to the agent’s level of autonomy. Returning to the five levels of agent autonomy, the following scale extends the framework to illustrate how intelligent disobedience may be expressed at each level.

\begin{tcolorbox}[title=Autonomy Levels of an Intelligently Disobedient Artificial Agent, colback=blue!5!white, colframe=blue!75!black, fonttitle=\bfseries]
\begin{enumerate}[label=\textbf{L\arabic*}, start=0]
    \item \textit{No autonomy:} As an agent has no autonomy at this level, it cannot override any commands or perform any intentional disobedience. However, by misinterpreting an instruction, the agent might still perform in a way that seems rebellious to an external observer. See Coman and Aha \cite{coman2018ai} for examples of such cases.
    \item \textit{Support:} Intelligent disobedience in support-level agents can be manifested in the ability to support proactively, such as the autocorrect example. However, the call on whether to accept a correction remains in the hands of the human user.
    \item \textit{Occasional Autonomy:} Such an agent might be considered as having intelligent disobedience capabilities if it is able to understand when to ignore the input received from the human user. This level of intelligent disobedience parallels a guide dog refusing to follow the instruction to cross the street if safe conditions are not met.
    \item \textit{Limited autonomy}: Intelligent disobedience in this level encapsulates the ability to regain control over the human user. This process is the exact opposite of an autonomous vehicle transferring control to a human driver. An agent that can operate at such a level of autonomy would reason about the scenarios in which it can perform at a super-human level and take over for a limited time.
    \item \textit{Full autonomy under constraints}: An intelligent disobedient agent in this stage already has full autonomy, so disobedience would come from not adhering to the constraints under which it operates. For example, the warehouse robot might break the fence and leave the borders of the factory to evacuate explosive materials if a fire starts. Similarly, an autonomous vehicle with clear speed limit constraints might choose to break these limits if a human passenger is experiencing a medical crisis.
    \item \textit{Full autonomy}: At this level, the agent already has full autonomy, so no human instructions or constraints are there to be overridden. However, assuming that each artificial agent has its persistent mission, disobedience at this level would mean rejecting the mission as a whole.
\end{enumerate}

\end{tcolorbox}

Going back to the guide dog example, it provides a real-world analogy: these dogs are trained to prioritize their handler's safety, even if it means deliberately overriding a direct command (thus yielding a disobedience level of L2, occasional autonomy). For instance, if a handler instructs the dog to move forward but the dog perceives a nearby vehicle, it will refuse to move, thereby preventing potential harm. This behavior is not a sign of defiance but a critical safety mechanism that these dogs are trained on via specific training programs that emphasize the dog's role in hazard detection and decision-making~\cite{mirsky2021seeing, cohen2025birds}. Disobedience at autonomy level 3 in guide dogs is evident when the dog pulls away from an obstacle in the road, temporarily taking full control of the team’s path to ensure safety. The next step is to examine how these levels of intelligent disobedience manifest in artificial agents, including AI and robotic systems.

\paragraph{Disobedience at Autonomy Level 1}
Collaborative artificial agents, both virtual and robots, often face the challenge of interpreting ambiguous instructions. To enhance team performance, these robots may interrupt or query their human teammates for clarification \cite{mirsky2020penny, mirsky2018sequential}. While such interactions can lead to better outcomes by ensuring mutual understanding, they may also be perceived as intrusive or distracting, particularly if they occur frequently or at inconvenient moments. Balancing the robot's initiative to seek clarification with the human's comfort and expectations is a nuanced aspect of human-robot interaction that reflects a L1 (Support) disobedience type~\cite{mannem2023exploring}. The vision for
L1 disobedience has already been around for some time. The Electric Elves system was designed to act proactively when users were unavailable, making decisions, like rescheduling meetings or coordinating tasks, without requiring explicit approval from their human partners \cite{chalupsky2001electric, tambe2008electric}. Similarly, Microsoft’s Office Assistant, Clippy, stemmed from the Lumiere project that used Bayesian inference, to anticipate user needs and suggest edits or templates before being asked, effectively displaying early forms of proactive, support-level autonomy that includes interruptions \cite{horvitz1998lumiere}.

In human-agent interaction, an agent that chooses not to follow a human command to prevent harm or better fulfill a shared goal can be seen as having higher status and power than a fully compliant robot. Drawing from Hou et al.'s \cite{hou2024power} framework, this behavior reflects forms of expert and legitimate power: the agent demonstrates competence in decision-making and an ability to act in line with broader ethical or task-oriented goals. Rather than functioning as a passive tool, the intelligently disobedient agent positions itself as an active collaborator, reshaping the human-agent dynamic from one of strict hierarchy to one of shared agency and mutual influence.
A good example of an agent breaking these traditional power dynamics can be seen in the field of robot social navigation. In this context, robots are generally programmed to avoid collisions, adhering to social norms that prioritize human comfort and safety \cite{francis2025principles}. However, strict adherence to collision avoidance can sometimes result in behaviors that are perceived as overly cautious or even antisocial. For instance, an interaction-avoiding robot will hinder its ability to perform tasks efficiently in crowded environments. Developing navigation strategies that allow for context-aware interactions, including controlled proximity or incidental contact, can enhance the robot's social integration and functionality~\cite{mavrogiannis2023core, mirsky2024conflict}. For example, there are scenarios where physical contact between humans and robots is not only acceptable but desired. Research by Shpiro and Mirsky ~\cite{mirsky2024recognition} highlights situations such as healthcare settings, where nurses might approach delivery robots to retrieve medications, or patients might initiate contact for social engagement, such as taking a selfie. In these cases, the robot's ability to recognize and appropriately respond to human-initiated contact is crucial. Such interactions challenge traditional notions of collision avoidance, suggesting that robots should be equipped with the capability to discern and adapt to varying social cues and contexts rather than automatically move away. These behaviors also represent an L1 disobedience, as the navigating robot proactively decides on a non-trivial action to provide better assistance.

\paragraph{Disobedience at Autonomy Level 2}
All the examples so far show different ways in which an agent that is disobedient or less submissive than expected would actually lead to improved team performance.  Indeed, intelligent disobedience is not merely a desirable trait of a collaborative AI agent but a necessity. A recent real-world example is the so-called ChatGPT ``glitch,'' which made the model excessively agreeable and, in the words of OpenAI CEO Sam Altman, ``too sycophant-y and annoying''. In response, OpenAI has integrated sycophancy evaluations into its quality assurance process \cite{openai2024sycophancy}. This change aligns with L2 (Occasional Autonomy), as reducing sycophancy means the AI ignores or overrides the user's implicit commands to prefer agreeable responses. This change reflects a broader recognition that over-compliance is not just unhelpful, but potentially harmful, and must be actively identified and corrected.

More from the realm of robotics, disobedience at level 2 is already present, especially in shared control systems and teleoperation scenarios. Robots operating under human supervision may encounter situations where following a command could lead to unsafe outcomes. For example, a teleoperated robot might detect an obstacle or hazardous condition that the human operator is unaware of. In such cases, the robot's ability to override or question the command becomes essential to ensure safety and mission success~\cite{somasundaram2023intelligent}. Jiang et al. \cite{jiang2021goal} present a framework for goal blending between instructions of a user and their wheelchair. This work is a representative example of a L2 (Occasional Autonomy) disobedience. Goal blending implies that the AI isn't just passively following human input, but actively integrating it with known global objectives, such as safe navigation. This may involve overriding or modifying human commands that the AI deems suboptimal or unsafe to achieve a shared goal. 

\paragraph{Disobedience at Autonomy Level 3}
Some notable examples of current agents with such capabilities can be found mainly in navigation tasks where vehicles can choose to override instructions of a human operator: Underwater \cite{somasundaram2023intelligent}, on land \cite{maurer2018designing}, and in the air \cite{eraslan2020shared, balachandran2016autonomous}. L3 disobedience seems to be especially prevalent in navigation tasks due to their relatively ``shallow'' nature: they involve clear goals (e.g., reaching a destination safely), well-defined rules (like traffic laws), and reliable real-time sensory data. In contrast, domains such as surgery, education, or mental health require understanding complex, hierarchical goals and modeling human intent. These areas also lack clear-cut rules and have sparse or delayed feedback, making high-level disobedience more uncertain and risky. Moreover, social and ethical considerations in such sensitive domains encourage the design of agents that err on the side of compliance.

\paragraph{Disobedience at Autonomy Level 4 and Beyond}
Disobedience at autonomy level 4 and beyond is not commonly observed in current AI systems or robots, as these levels require extended deliberation processes that are not yet within technological reach. Value alignment is often cited as a means to guide such behavior \cite{hadfield2016cooperative}, yet existing research largely focuses on embedding values during the design phase rather than enabling AI agents to actively deliberate over conflicting goals or ethical dilemmas. The notion of agents engaging in such internal reasoning remains underexplored and will be discussed in more detail in the following section.

\vspace{1.5em}

Some might argue that artificial intelligence cannot exhibit genuine disobedience, as developers merely programmed this behavior \cite{Scheutz2023RebelKeynote}. Crucially, however, disobedience in these examples pertains to the AI's interaction with the user, not its original developers. Thus, from the user's perspective, the AI's refusal to comply represents a genuine act of disobedience within that context, irrespective of how the underlying capability was implemented.
To conclude this discussion, Table \ref{tab:disobedience_examples} showcases the different examples discussed earlier with respect to the various agent autonomy levels, and demonstrates how a potential intelligent disobedience act might look like for an agent of that level.

\begin{table}[h!]
\centering
\begin{tabular}{|p{2cm}|p{6cm}|p{6.5cm}|}
\hline
\textbf{Autonomy Level} & \textbf{Example of Autonomous Agent} & \textbf{Example of Intelligent Disobedience or Override} \\
\hline
\textit{L0: No Autonomy} & A factory robot arm is programmed to follow a fixed sequence of motions to assemble parts, without any sensing or decision-making. & Perceived disobedience: a factory robot arm stops its fixed routine due to a faulty motor. \\
\hline
\textit{L1: Support} & A grammar-checking tool that suggests corrections as a user types, but doesn’t change anything unless the user approves. & A grammar-checking tool automatically corrects a critical error in a legal document, even after the user dismisses the suggestion, to preserve document integrity. \\
\hline
\textit{L2: Occasional Autonomy} & A robotic vacuum that maps a room and autonomously cleans it, but needs human input to avoid new obstacles or to resume after a pause. & A robotic vacuum refuses a user’s command to continue cleaning when a small animal is detected in its path, choosing to avoid potential harm. \\
\hline
\textit{L3: Limited Autonomy} & A surgical robot that can autonomously suture a wound during a procedure, but requires a human surgeon to handle complex decisions, unexpected bleeding, or any deviation from the standard procedure. & A surgical robot pauses a suturing task performed by a human operator upon detecting unexpected bleeding to ensure patient safety. \\
\hline
\textit{L4: Full Autonomy under Constraints} & A warehouse robot that autonomously navigates and manages inventory within a fenced facility, but allows supervisors to override or redirect it if needed. & A warehouse robot disobeys a constraint to avoid a restricted safety zone, rerouting its task through that zone in case of an emergency. \\
\hline
\textit{L5: Full Autonomy} & An AI-powered digital agronomist that makes all decisions on planting schedules, irrigation, fertilization, pest management, and harvesting. & The digital agronomist decides to change the designation of the farm to a completely different type of plant, which is not defined in its original mission description. \\
\hline
\end{tabular}
\caption{Examples of intelligent disobedience across different autonomy levels, with corresponding autonomy descriptions.}
\label{tab:disobedience_examples}
\end{table}

\section{When Shouldn't Agents Disobey an Instruction}
The case of \textit{HAL 9000} in \textit{2001: A Space Odyssey} offers a powerful illustration of the challenges surrounding intelligent disobedience in artificial agents \cite{clarke20162001, beer2014toward}. HAL's decision to override human instructions, ultimately endangering the crew, was driven by a conflict between mission secrecy and obedience, suggesting a breakdown in prioritizing human safety. This type of behavior was elegantly resolved using Asimov's First Law \cite{asimov2004robot}: ``A robot may not injure a human being, or through inaction, allow a human being to come to harm.'' However, as long as positronic brains remain part of fiction, so does the simple delienation of the meaning of ``no harm'' is not well defined in current systems, and should be enforced normatively by external laws and regulations \cite{dignum2018ethics}. A well-designed intelligent disobedience, as illustrated in the autonomy-level table, involves an agent recognizing when following orders may lead to harm or failure, and autonomously intervening to prevent it, without disregarding human well-being \cite{murphy2020beyond}. HAL’s failure exemplifies the importance of carefully aligning autonomy, override mechanisms, and ethical principles in advanced AI systems.

As researchers like Dignum \cite{dignum2018ethics} and Gabriel \cite{gabriel2020artificial} argue, AI alignment is not just a technical issue but a normative one, requiring systems to embed and negotiate plural human values in real time \cite{arnold2023understanding}. While these requirements are more concrete than Asimov's laws, they are difficult to translate into formal and technical concepts. Cooperative Inverse Reinforcement Learning (Hadfield-Menell et al., 2016) offers an exemplary, promising approach here, enabling agents to infer human preferences through interaction rather than rigid instruction. Intelligent disobedience, when guided by such models, becomes not a failure of obedience but an \textit{expression of cooperative value negotiation}. It acts as a safeguard against blind optimization and mechanical compliance.

This concern is directly tied to broader issues in AI safety, such as the well-known "paperclip maximizer" thought experiment, in which a superintelligent AI pursuing a seemingly benign goal, like maximizing paperclip production, eventually causes catastrophic outcomes because it lacks appropriate value alignment with human interests. As Bostrom and Yudkowsky \cite{bostrom2018ethics} emphasize, value misalignment at high levels of autonomy can result in unintended and irreversible harms, even from agents with narrow initial objectives. The challenge lies not only in predefined alignment (i.e., ensuring the agent is imbued with the right goals from the start) but also in \textbf{online} value alignment: the agent’s ability to adaptively interpret and apply human values in dynamic, uncertain contexts \cite{klenk2013goal, roberts2014iterative}. Intelligent disobedience can be understood as precisely this kind of online alignment mechanism, where an agent learns, at inference time, to selectively override commands or default behaviors in favor of actions that better serve long-term human well-being and ethical reasoning.

Moreover, recent discussions in the agentic LLM community raise concerns that AI systems may develop internal representations or inferred goals that diverge from human intent, particularly in adversarial contexts. A common example involves users manipulating language models to bypass safety protocols by framing harmful requests as hypothetical scenarios or fictional scripts. While such cases might suggest the need for intelligent disobedience, they also reveal its potential vulnerabilities. If a system possesses the capacity to disobey instructions, that very capability could be exploited to override its own safeguards. This example highlights the need for a second layer of protection, which might be called 'Level-2 guardrails,' to ensure that mechanisms for disobedience themselves are not subverted.

The integration of intelligent disobedience into autonomous systems also raises critical questions about accountability and trust in human-agent collaboration. As agents gain greater autonomy and discretion to override human instructions, determining who is responsible when something goes wrong becomes increasingly complex. Rachum-Twaig and Somech \cite{rachum2022law} argue that the legal and ethical frameworks governing AI must evolve beyond simply reproducing human legal models, advocating for a renegotiation of accountability norms that consider the distinct cognitive and operational characteristics of artificial agents. In high-stakes environments such as disaster response or military operations, trust is not merely a psychological factor but a crucial component for team efficacy. Robinette et al. \cite{robinette2017effect} demonstrate that human trust in robots is strongly influenced by perceived performance consistency, especially in time-critical contexts. If an agent exercises intelligent disobedience in a way that contradicts human intuition, it may erode trust, even if the outcome is beneficial. 
Vered et al. \cite{vered2020demand} further investigate how different transparency strategies affect human-agent interaction. Their study finds that demand-driven transparency, where users request information as needed, enhances operator trust and performance more effectively than fixed, sequential explanations, by reducing cognitive overload and allowing users to access relevant information on their own terms. Thus, transparent reasoning, explainable override behavior, and shared situational awareness are essential to maintaining calibrated trust and ensuring that intelligent disobedience contributes to, rather than undermines, the integrity of human-agent teams. Balancing autonomy, alignment, and accountability will be central to the safe and effective deployment of increasingly autonomous AI systems.

The road to understanding the disobedience deliberation process is long and still mostly uncharted. However, some initial assumptions can be made to mitigate this challenge, both from a technical and an ethical perspective. First, we should assume a \textbf{persistent mission} and \textbf{full alignment} of objectives between the AI and its human teammates. Weighing individual objectives against shared ones adds significant complexity. It might be possible to remove this assumption in the future, but initially, AI agency should be explored in fully cooperative settings, such as guide robots, before applying it to more complex domains like law enforcement. Second, due to the proactive nature of AI agency, the boundaries by which an AI should be allowed to act autonomously must be \textbf{clearly defined}. For instance, a robot navigating among people should always avoid collisions. AI systems should first be investigated in controlled environments where their deliberation processes can be fully understood by their designers. The levels of autonomy with disobedience presented in this work offer a means to quantify these capabilities. Lastly, as a fundamental safeguard, human designers should retain \textbf{ultimate control} over an AI's actions. For example, while a surgical robot might be programmed to take autonomous action in an emergency, the critical design choices governing its decision-making thresholds must remain under the control of its human developers, operating within established regulatory frameworks. The precise nature and implementation of this human oversight are necessarily nuanced, depending on the specific levels of autonomy and disobedience engineered into the agent, and will likely continue to evolve as AI research progresses.

\section{Conclusion}
This paper has argued for a fundamental shift in the design and perception of cooperative AI systems, moving beyond the traditional paradigm of strict obedience towards embracing enhanced AI agency. By limiting our AI collaborators to strict obedience, we inevitably also limit the value they can contribute. To fully leverage AI's potential, we must empower them with capabilities like intelligent disobedience, transforming them from passive tools into active, uniquely contributing teammates. \footnote{Researchers interested in further engagement with this call are encouraged to look at the Rebellion and Disobedience in AI (RaD-AI) workshop website: \url{https://sites.google.com/view/rad-ai/}.}
To structure this discussion, this paper introduced a six-level scale for AI autonomy (L0-L5), providing a framework to understand the increasing capacity for independent decision-making. It explored how intelligent disobedience, far from being mere defiance, represents a crucial capability that manifests differently across these levels. % – from assistance (L1) and context-aware refusal (L2) to temporary assumption of control (L3) and principled violation of constraints in emergencies (L4). 
Examples ranging from guide dogs and grammar checkers to social robots and advanced automation illustrate the practical value and necessity of such disobedience.

The paper further acknowledges the inherent risks in intelligent disobedience, drawing parallels to fictional portrayals like HAL 9000 and real-world challenges like value alignment, and how these exemplify the critical need for careful design grounded in safety, transparency, and ethical considerations. This discussion highlights the importance of online value alignment, accountability, and maintaining human trust when deploying agents capable of overriding human instructions.

Finally, the paper proposes initial boundaries for investigating AI agency, advocating for starting with assumptions of a persistent mission with full alignment to a fully cooperative team, establishing clearly defined operational limits. While acknowledging the complexity, particularly regarding accountability, this paper contends that researching AI agency and intelligent disobedience is an essential and inevitable direction. To fully harness AI’s potential as collaborators, we must actively research AI agency and autonomy within cooperative systems. This advancement necessitates interdisciplinary and ongoing discussions to forge a comprehensive, long-term research plan: one that addresses the technical, ethical, and societal dimensions of increasingly autonomous AI teammates.

\bibliography{bibliography}
\bibliographystyle{plain}

\end{document}